\documentclass[letterpaper]{article}
\usepackage[preprint]{aaai2027}
\usepackage[hyphens]{url}
\usepackage{graphicx}
\usepackage{natbib}
\usepackage{caption}
\usepackage{algorithm}
\usepackage{algorithmic}
\usepackage{booktabs}
\usepackage{multirow}
\usepackage{makecell}
\usepackage{pifont}
\definecolor{tabhead}{gray}{0.92}
\definecolor{oursrow}{gray}{0.93}
\newcommand{\yes}{\ding{51}}
\newcommand{\no}{\ding{55}}
\newcommand{\mlogo}[1]{\raisebox{-0.26\height}{\includegraphics[height=11pt]{Figures/logos/#1}}}
\usepackage{colortbl}
\newcommand{\sparo}{\textsc{Sparo}}
\title{Beyond Prompt or Skill? Attribution-Guided Optimization of Modular LLM Programs}
\author{
    \textbf{Haoran Shou}\textsuperscript{1,*}
    \quad
    \textbf{Haoyue Liu}\textsuperscript{1,2,*}
    \quad
    \textbf{Yu Huo}\textsuperscript{1}
    \quad
    \textbf{Kun Zeng}\textsuperscript{3}
    \quad
    \textbf{Xiaoying Tang}\textsuperscript{1,2,\ensuremath{\dagger}}
}

\affiliations{
    \textsuperscript{1}
    \textnormal{School of Science and Engineering,
    The Chinese University of Hong Kong, Shenzhen 518172, China}
    \\[0.3em]
    \textsuperscript{2}
    \textnormal{Shenzhen Future Network of Intelligence Institute (FNii-Shenzhen)}
    \\[0.3em]
    \textsuperscript{3}
    \textnormal{Sun Yat-sen University}
    \\[0.5em]
    \textsuperscript{*}\textnormal{Equal contribution.}
    \quad
    \textsuperscript{\ensuremath{\dagger}}\textnormal{Corresponding author.}
}
\begin{document}

\maketitle

\begin{abstract}
Large language models can solve increasingly diverse reasoning tasks, yet their performance remains highly sensitive to task prompts, intermediate instructions, and the way reusable problem-solving knowledge is incorporated. Existing optimization methods usually focus on only one part of this design space: they either optimize a monolithic prompt, or separately induce and refine skills from model traces. As a result, they lack a principled mechanism for deciding which component should be updated when failures occur, and they rarely optimize prompts, skills, and skill-use policies in a unified framework. We propose \sparo{} (\textbf{S}kill, \textbf{P}rompt, \textbf{A}nd \textbf{R}outing \textbf{O}ptimization), a framework that jointly optimizes task instructions, reusable skill blocks, and routing rules. It performs controlled counterfactual evaluations, converts examples' effects into a probabilistic responsibility distribution over prompt, skill, and routing components, samples one component from that distribution, and applies the corresponding targeted mutation. This design moves language-program optimization beyond global prompt rewriting toward structured, reusable, and selectively activated task knowledge. Across five benchmarks and five worker models, \sparo{} consistently outperforms both prompt-centered and skill-centered optimization baselines. These results suggest that effective language-program optimization depends not only on discovering useful task knowledge, but also on deciding where that knowledge should be stored and when it should be activated.
\end{abstract}

\begin{figure}[!t]
\centering
\includegraphics[width=0.45\textwidth]{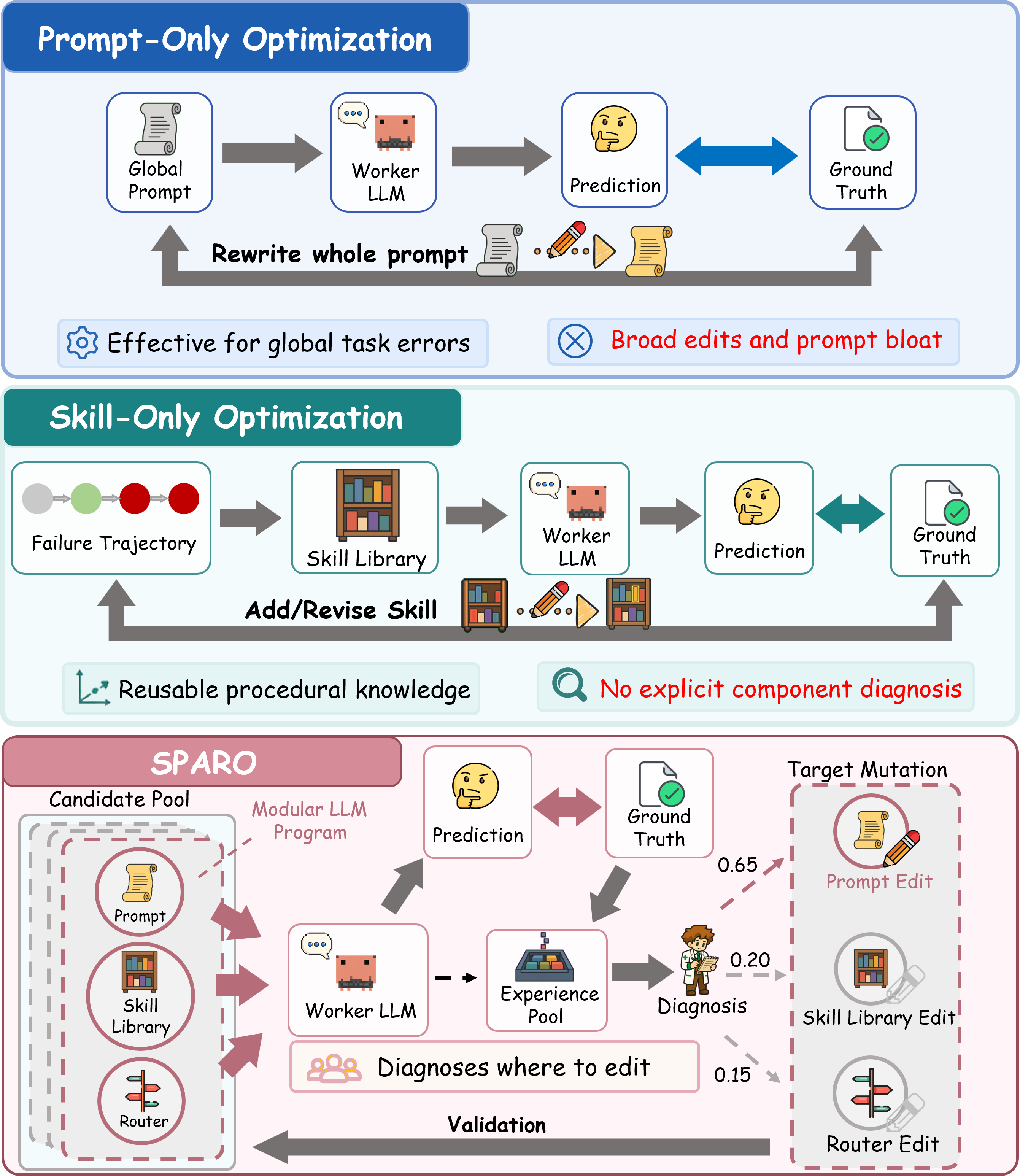}
\caption{Overview of language-program optimization paradigms.
Prompt-only optimization rewrites the global prompt for any failure;
skill-only optimization distills reusable procedures but never diagnoses
which component is at fault; \sparo{} treats the program as a modular object
$P=(p,\mathcal{B},r,t)$ and edits only the component that its attribution
step samples.}
\label{fig:overview}
\end{figure}

\section{Introduction}

Adapting large language models (LLMs) to downstream tasks increasingly
involves optimizing the language program around a fixed model rather
than updating the model parameters themselves
\citep{brown2020language,ouyang2022training}. A practical language
program may combine a global task instruction, formatting constraints,
examples, decomposition rules, and reusable natural-language
procedures~\citep{dspy2024,opro2023,textgrad2024,gepa2025}. However, its modularity also creates a credit-assignment problem:
end-to-end task feedback does not directly identify which program
component should be revised \citep{opsahlong2024mipro}.
A failed prediction may result from an underspecified prompt, missing
or harmful procedural knowledge, or an incorrect decision about when that
knowledge should be activated.

Existing language-program optimizers typically search only one part of
this design space. Prompt-centered methods improve instructions or
prompt programs through iterative generation, textual feedback, or
reflective mutation
\citep{opro2023,protegi2023,textgrad2024,gepa2025}. They are effective
when a failure can be repaired through global instruction rewriting, but
may encode recurring local exceptions directly into the prompt.
Skill-oriented methods instead extract or refine reusable procedures
from trajectories and agent experience
\citep{trace2skill2026,skillopt2026,skillgen2026}. These approaches make
reusable knowledge explicit, but generally do not decide whether the
next update should revise a skill, change the global prompt, or modify
the mechanism that routes skills to inputs. Figure~\ref{fig:overview}
contrasts these two families with the approach we propose.

This paper asks: \emph{can an optimizer determine which component of a
modular LLM program should change before generating the change?}
Answering this question requires attribution rather than reflection
alone. It also requires evidence beyond the current minibatch: an
isolated error may not reveal whether a failure mode is recurring, while
historical examples may change from failures to successes as the program
evolves~\citep{shinn2023reflexion,zhao2024expel}. We therefore maintain a cluster-balanced cross-round experience
pool and relabel its examples under the current program before each
optimization step.

We introduce \sparo{}, an attribution-guided optimizer for modular LLM
programs. \sparo{} represents a program as a global prompt, a reusable skill
library, a semantic router, and an integration template. At each
optimization step, \sparo{} retrieves diverse historical evidence, evaluates
controlled counterfactual program variants, and converts their effects
into a probabilistic responsibility distribution over the prompt, skill
library, and router. It then samples a component from this distribution
and applies the corresponding mutation. The resulting candidate is
evaluated on the validation set and maintained using Pareto-aware
program-pool selection.

We evaluate \sparo{} on five heterogeneous reasoning and question-answering
benchmarks using five instruction-tuned and multimodal worker models.
\sparo{} obtains the highest score in every evaluated model--task
setting, improving over both prompt-centered and skill-centered
optimization methods. The results indicate that the
benefit does not arise merely from adding more textual instructions or
more skills, but from deciding where newly discovered task knowledge
should be represented and when it should be activated.

We summarize our contributions as follows:
\begin{itemize}
    \item We formulate modular language-program optimization as a
    component credit-assignment problem over global prompts, reusable
    skills, and routing mechanisms.

    \item We introduce a cluster-balanced cross-round experience pool
    that preserves diverse historical evidence and dynamically relabels
    examples under the current program.

    \item We propose probabilistic counterfactual attribution and
    targeted component mutation, enabling the optimizer to sample the
    program component to update from a responsibility distribution
    induced by current evidence.
\end{itemize}

\section{Related Work}

\paragraph{Black-box prompt optimization.}
Black-box prompt optimization searches over natural-language
instructions while keeping model parameters fixed. APE generates and
selects candidate instructions using task-level scores
\citep{zhou2023large}, OPRO proposes new prompts from previously
evaluated prompt--score pairs \citep{opro2023}, and ProTeGi and TextGrad
use natural-language feedback as gradient-like optimization signals
\citep{protegi2023,textgrad2024}. Self-Refine studies iterative revision
through self-generated feedback \citep{madaan2023selfrefine}, while
PromptBreeder and GEPA maintain and evolve populations of prompt
candidates \citep{fernando2024promptbreeder,gepa2025}. These methods
mainly optimize prompts or prompt programs, without deciding whether a
failure should instead be addressed through reusable skills or routing.

\paragraph{Skill learning and reusable procedures.}
Language agents can improve by retaining feedback, trajectories, or
reusable procedures. Reflexion stores verbal feedback in episodic memory
\citep{shinn2023reflexion}, ExpeL extracts transferable knowledge from
past experience \citep{zhao2024expel}, and Voyager maintains a growing
library of reusable skills \citep{wang2023voyager}. Recent methods such
as Trace2Skill, SkillOpt, and SkillGen further distill, evolve, or verify
agent skills \citep{trace2skill2026,skillopt2026,skillgen2026}. However,
skill learning alone does not determine whether an error originates from
missing knowledge, incorrect skill content, the global prompt, or skill
activation.

\paragraph{Structured language programs and routing.}
Structured language-program frameworks represent applications as
compositions of multiple modules. DSPy optimizes instructions and
demonstrations in declarative LM pipelines \citep{dspy2024}, while MIPRO
addresses optimization and credit assignment in multi-stage language
programs \citep{opsahlong2024mipro}. aPSF factorizes a prompt into separately optimized factors and scores them
by single-factor intervention \citep{apsf2026}, though its factors are all
parts of a single prompt rendered for every input. Input-dependent
component selection is also related to sparse expert routing
\citep{fedus2022switch} and learned tool invocation
\citep{schick2023toolformer}. In contrast, \sparo{} assigns credit across
heterogeneous components of a single modular program.

\paragraph{Component attribution in \sparo{}.}
\sparo{} differs from structured prompt-program optimization in what it
attributes over: the components are heterogeneous rather than slots of one
prompt, the skill library persists across inputs, and the router decides at
inference time which skills an input sees, making missed and harmful
activations their own counterfactuals. Candidates are retained by GEPA's
validation-based Pareto mechanism \citep{gepa2025}; our contribution is
attribution-guided candidate generation, not a new selection rule.

\section{\sparo{}: Attribution-Guided Modular Optimization}

\subsection{Problem Setup}

\sparo{} optimizes a modular language program while keeping the worker model fixed. Each task example is a pair $(x,y)$ with a metric $M(\hat{y},y)$, and the optimizer can query the worker LLM but cannot update its parameters. This black-box setting follows recent language-program optimization work, where improvement comes from rewriting textual programs rather than training model weights \citep{opro2023,textgrad2024,gepa2025}.

\begin{figure*}[!htbp]
\centering
\includegraphics[width=0.95\textwidth]{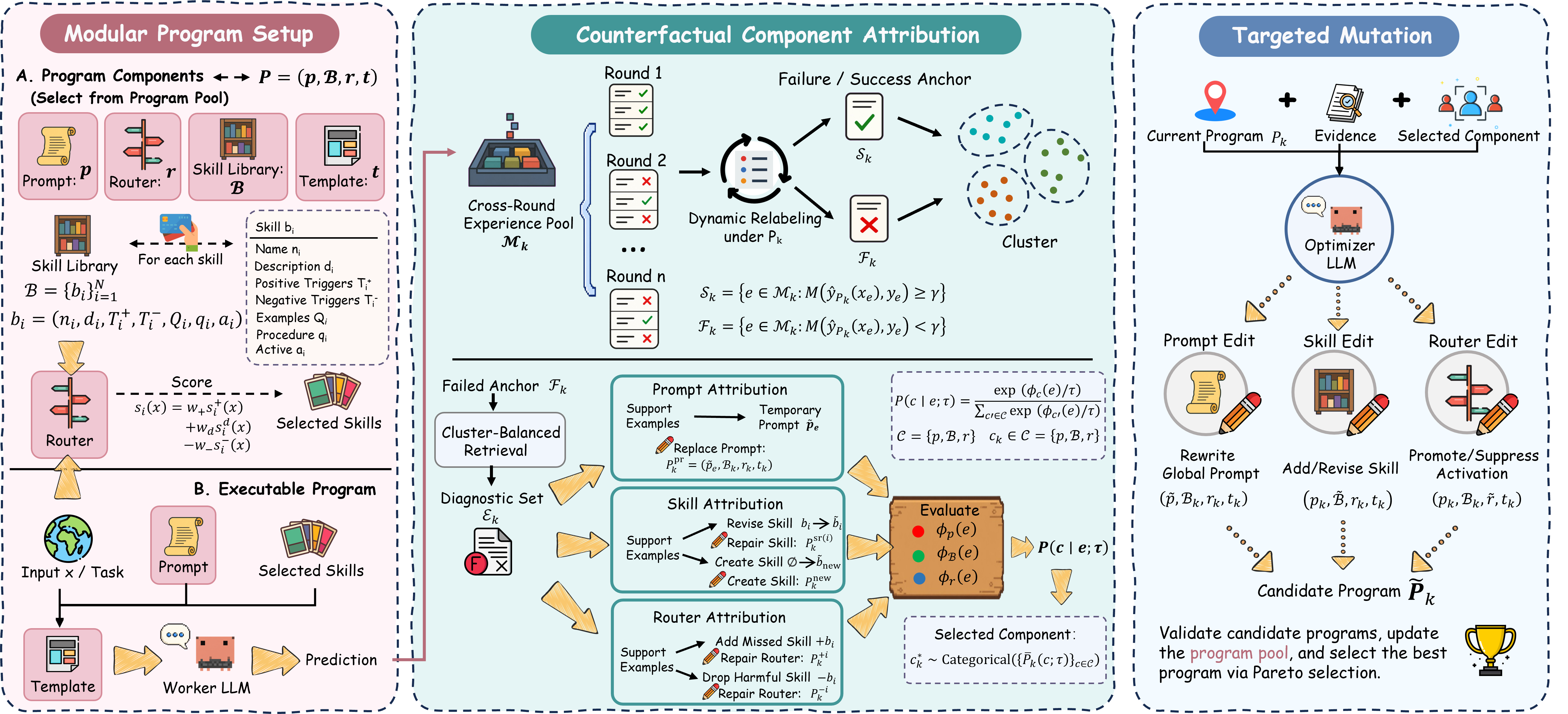}
\caption{Overview of \sparo{}. \sparo{} maintains a cross-round evidence pool
and optimizes a modular LLM program by performing counterfactual component
attribution over the prompt, skill library, and router, then sampling a
component from that responsibility distribution and applying the
corresponding targeted mutation.}
\label{fig:pipeline}
\end{figure*}

Following modular language-program frameworks
\citep{dspy2024,opsahlong2024mipro}, we represent a program as $P=(p,\mathcal{B},r,t)$, where $p$ is the main prompt, $\mathcal{B}$ is a library of skill blocks, $r$ is a routing rule, and $t$ is an integration template (Figure~\ref{fig:pipeline}, left). Given input $x$, the program renders a final instruction for the worker model using the selected skill set $\mathcal{B}_x$ and predicts
\[
\hat{y}_P(x)=f_{\theta}\bigl(t(x,p,\mathcal{B}_x)\bigr).
\]

With the template $t$ held fixed, the objective is to find the best program within a fixed optimization budget:
\[
\begin{array}{rcl}
P^\star
&=&
\arg\max_{P\in\mathcal{P}} J_{\mathrm{val}}(P),
\\[6pt]
J_{\mathrm{val}}(P)
&=&
\frac{1}{|\mathcal{D}_{\mathrm{val}}|}
\sum_{(x,y)\in\mathcal{D}_{\mathrm{val}}}
M(\hat{y}_P(x),y).
\end{array}
\]

\subsection{Modular Program Composition}

\sparo{} treats an LLM program as a collaboration between global instructions and selectively activated reusable knowledge. The main prompt $p$ is always present and defines the task contract: what the input means, what the answer should contain, and which global constraints such as output format or answer style must be obeyed. Since these instructions apply to all examples, modifying the main prompt has the broadest effect on program behavior.

The skill library $\mathcal{B}=\{b_i\}_{i=1}^{N}$ stores reusable local procedures. Inspired by persistent and
structured skill libraries
for retrieval and composition
\citep{wang2023voyager,zeng2026groupskillsgroupstructuredskill,
zeng2026sigmaskillincidencegraphscompositional}, \sparo{} represents each
skill as
\[
b_i=(n_i,d_i,T_i^{+},T_i^{-},Q_i,q_i,a_i),
\]
where $n_i$ is the skill name, $d_i$ describes its applicability, $T_i^{+}$ and $T_i^{-}$ are positive and negative trigger sets, $Q_i$ contains representative queries, $q_i$ is the natural-language procedure, and $a_i\in\{0,1\}$ indicates whether the skill is active. The routing metadata specifies when a skill should or should not be used, while the procedure specifies how the worker model should handle the corresponding recurring subproblem.

The router $r$ selects skills by comparing the semantic representation of the current input with their routing metadata. Let $h(\cdot)$ be a fixed sentence embedding model~\citep{reimers2019sentencebert} and let $\mathbf{x}=h(x)$ denote the embedding of the routing view of input $x$. For each active skill, \sparo{} first computes the positive relevance score
\[
s_i^{+}(x)
=
\max_{u\in T_i^{+}\cup Q_i}
\cos(\mathbf{x},h(u)),
\]
which measures how closely the input matches at least one intended use case or representative query of the skill. It also computes the negative relevance score
\[
s_i^{-}(x)
=
\max_{v\in T_i^{-}}
\cos(\mathbf{x},h(v)),
\]
which measures how closely the input matches a known condition under which the skill should not be activated. We define $s_i^{-}(x)=0$ when $T_i^{-}=\emptyset$.

In addition to these trigger-level signals, \sparo{} measures the overall semantic relevance of the skill through
\[
s_i^{d}(x)=\cos(\mathbf{x},h(d_i)),
\]
where the description $d_i$ provides a broader summary of the skill's applicability beyond the finite set of triggers. The final routing score combines the three signals:
\[
s_i(x)
=
w_{+}s_i^{+}(x)
+
w_d s_i^{d}(x)
-
w_{-}s_i^{-}(x).
\]The weights $w_{+}$, $w_d$, and $w_{-}$ control the relative influence of these complementary signals.

Conceptually related to sparse expert routing
\citep{fedus2022switch}, the router selects at most $K_r$ active skills. The integration template $t$ then combines the input, the main prompt, and the selected skill procedures.

This decomposition defines the action space of \sparo{}: a prompt edit changes the global task contract, a skill edit changes reusable procedural knowledge, and a routing edit modifies the metadata that determines when a skill is selected. The remaining subsections describe how \sparo{} collects evidence, attributes failures to these components, and verifies candidate edits.

\subsection{Cluster-Balanced Evidence Retrieval}

\sparo{} maintains a cross-round experience pool
$\mathcal{M}_k$ to provide the optimizer with diverse historical
evidence. A single optimization minibatch often contains only a few
examples of a recurring failure pattern, making it difficult to decide
whether the problem should be addressed by a prompt edit, a skill
change, or a routing modification.

Unlike a static failure memory, \sparo{} stores execution traces, model
outputs, feedback, and examples without assigning permanent labels.
At each optimization step, the pool is re-evaluated using the current
program $P_k$ and split into failure and success sets by a fixed
success threshold $\gamma$ on the task metric:

\[
\mathcal{F}_k=
\{e\in\mathcal{M}_k:
M(\hat y_{P_k}(x_e),y_e)<\gamma\},
\]

\[
\mathcal{S}_k=
\{e\in\mathcal{M}_k:
M(\hat y_{P_k}(x_e),y_e)\geq\gamma\}.
\]

The pool maintains diversity through clustering over input and error
features. During mutation generation, \sparo{} retrieves related failures
and successful examples from corresponding clusters, providing the
optimizer with consistent evidence rather than isolated mistakes.

\subsection{Counterfactual Attribution and Component Choice}

\sparo{} forms a distribution over which program component to revise before
generating a mutation. At optimization step $k$, it retrieves a set of examples $\mathcal{E}_k$.
For counterfactuals that require generating temporary prompt or skill
proposals, examples from each failure cluster are divided into support
and query subsets: the support examples generate the proposal, and the
held-out query examples measure its effect.

For each query example $e=(x,y)\in\mathcal{E}_k$, let
\[
s_0(e)=M(\hat{y}_{P_k}(x),y)
\]
denote the score of the current program. \sparo{} constructs
component-specific counterfactuals that modify one component while
keeping the others fixed.

\paragraph{Prompt attribution.}
The optimizer uses the corresponding support examples to generate a
temporary repaired prompt $\tilde{p}_e$. Let
$P_k^{\mathrm{pr}}=(\tilde{p}_e,\mathcal{B}_k,r_k,t_k)$ be the program obtained
by replacing only the global prompt. The prompt contribution is
\[
\phi_p(e)
=
\left[
M(\hat{y}_{P_k^{\mathrm{pr}}}(x),y)-s_0(e)
\right]_+,
\]
where $[z]_+=\max(z,0)$. A positive value indicates that revising the
global task specification improves the prediction.

\paragraph{Routing attribution.}
Let $\mathcal{B}_x\subseteq \mathcal{B}_k$ be the skills selected for input $x$, and let
\[
U_e=
\mathrm{TopKRelevant}
\left(x,\mathcal{B}_k\setminus \mathcal{B}_x;K_d\right)
\]
be the $K_d$ highest-scoring unselected skills under the router
relevance function. \sparo{} tests a missed activation by adding one skill
from $U_e$, and an incorrect activation by removing one skill from
$\mathcal{B}_x$:
\[
\begin{array}{rcl}
s_{\mathrm{add}}(e)
& = &
\displaystyle\max_{b_i\in U_e}
M\!\left(\hat{y}_{P_k^{+\!i}}(x),y\right),
\\[4pt]
s_{\mathrm{drop}}(e)
& = &
\displaystyle\max_{b_i\in \mathcal{B}_x}
M\!\left(\hat{y}_{P_k^{-\!i}}(x),y\right).
\end{array}
\]
Empty candidate sets are assigned the baseline score $s_0(e)$. The
routing contribution is
\[
\phi_r(e)
=
\max
\left\{
[s_{\mathrm{add}}(e)-s_0(e)]_+,
[s_{\mathrm{drop}}(e)-s_0(e)]_+
\right\}.
\]
The two terms respectively identify a useful existing skill that was
missed and an irrelevant or harmful skill that was incorrectly
activated.

\paragraph{Skill attribution.}
\sparo{} distinguishes incorrect existing skill content from missing
procedural knowledge. It considers
\[
R_e=
\mathrm{TopKRelevant}
\left(x,\mathcal{B}_x\cup U_e;K_d\right)
\]
as the candidate set for skill repair. For each $b_i\in R_e$, the
optimizer generates a temporary revised skill $\tilde{b}_i$, and
$P_k^{\mathrm{sr}(i)}$ replaces only that skill while preserving its
activation status. The best repair score is
\[
s_{\mathrm{sr}}(e)
=
\max_{b_i\in R_e}
M(\hat{y}_{P_k^{\mathrm{sr}(i)}}(x),y).
\]

The optimizer also proposes a temporary new skill
$\tilde{b}_{\mathrm{new}}$ from the support examples and directly
injects it into the selected skill set, producing
$P_k^{\mathrm{new}}$. The skill contribution is
{\fontsize{8.5pt}{8pt}\selectfont
\[
\phi_{\mathcal{B}}(e)
=
\max\left\{
\left[s_{\mathrm{sr}}(e)-s_0(e)\right]_+,
\left[
M(\hat{y}_{P_k^{\mathrm{new}}}(x),y)-s_0(e)
\right]_+
\right\}.
\]
}
The first term supports revising an existing skill, whereas the second
supports creating a new skill.

\paragraph{Cross-example aggregation.}
Let $C=\{p,\mathcal{B},r\}$ be the editable components; the integration
template $t$ is held fixed. \sparo{} converts the component contributions
into an example-level responsibility distribution:
{
\[
\begin{array}{rcl}
P(c\mid e;\tau)
&=&
\displaystyle
\frac{\exp\bigl(\phi_c(e)/\tau\bigr)}
{\sum_{c'\in C}\exp\bigl(\phi_{c'}(e)/\tau\bigr)},
\\[4pt]
\bar{P}_k(c;\tau)
&=&
\displaystyle
\frac{1}{|\mathcal{E}_k|}
\sum_{e\in\mathcal{E}_k}P(c\mid e;\tau),
\\[4pt]
c_k^\ast
&\sim&
\mathrm{Categorical}
\left(
\left\{
\bar{P}_k(c;\tau)
\right\}_{c\in C}
\right).
\end{array}
\]
}
As $\tau\to0$ each example casts a hard component vote, so $\bar{P}_k$
approaches the vote shares and selection stays stochastic; larger $\tau$
preserves attribution uncertainty.

\paragraph{Mutation selection.}
If $c_k^*=p$, \sparo{} applies \textsc{PromptEdit}. If $c_k^*=r$, it uses
the larger aggregate gain between skill addition and skill removal to
choose whether to promote or suppress activation. If $c_k^*=\mathcal{B}$, it
uses the larger aggregate gain between existing-skill repair and
new-skill proposal to choose between revising and creating a skill.
The resulting candidates are finally evaluated through
validation-based candidate-pool selection.

\subsection{Targeted Mutations from Relabeled Evidence}

After sampling $c_k^\ast$ from $\bar{P}_k$, \sparo{} generates the corresponding component-specific mutation. \textsc{PromptEdit} rewrites the global instruction when failures point to broad task interpretation or formatting errors. \textsc{SkillEdit} creates or revises reusable blocks when a cluster of failures shares a local reasoning pattern. \textsc{RoutingEdit} changes triggers or routing rules when useful skills exist but are selected incorrectly.

The cluster-balanced pool determines the evidence shown to the optimizer LLM, which is broader than the attribution set $\mathcal{E}_k$. \sparo{} first retrieves pool examples from clusters that are similar to the current failures. Within each relevant cluster, examples are ranked by similarity and recency; across clusters, \sparo{} samples in a balanced order. This gives the optimizer multiple same-type errors rather than a single failing example, making the proposed edit more likely to capture a reusable pattern, while success anchors discourage edits that break already-solved cases.

\subsection{Candidate Evaluation and Program Pool Selection}

After applying a component-specific mutation $m$ to the selected
parent program $P_k$, \sparo{} obtains a candidate program
$\widetilde{P}_k=m(P_k)$. We evaluate and maintain candidate programs
using the validation-based program-pool and instance-level Pareto
selection mechanism of GEPA~\citep{gepa2025}. Specifically, each
candidate is evaluated on the validation set and retained if it
contributes to the Pareto frontier or satisfies the pool's
quality--diversity criterion. Finally, the program with the highest validation score is chosen as the returned program.

\section{Experiments}

\subsection{Experimental Setup}

\textbf{Tasks and datasets.} We select five benchmarks that vary along modality, reasoning structure, and the type of language-program adaptation they reward. FinQA requires numerical reasoning over financial reports and tables \citep{chen2021finqa}; CLUTRR tests the induction and reuse of relational rules \citep{sinha2019clutrr}; ChartQA combines visual chart understanding with answer-format control \citep{masry2022chartqa}; ProofWriter evaluates multi-step deductive reasoning \citep{tafjord2021proofwriter}; and SearchQA requires concise answer extraction from noisy search snippets \citep{dunn2017searchqa}. 

\textbf{Baselines.} We compare \sparo{} with six recent language-program optimizers covering two complementary families. Prompt-centered methods include OPRO \citep{opro2023}, TextGrad \citep{textgrad2024}, GEPA \citep{gepa2025}, and the factorized prompt optimizer aPSF \citep{apsf2026}. Skill-centered methods include Trace2Skill \citep{trace2skill2026} and SkillOpt \citep{skillopt2026}. The comparison therefore tests whether jointly selecting and optimizing prompts, skills, and routing improves over either global prompt search or reusable-skill learning alone.

\textbf{Models and protocol.} Worker models span Qwen3.5-4B, GLM-4.6V-Flash, InternVL3.5-8B, MiniCPM-V-4.5, and a relatively larger model Gemma4-31B~\citep{qwen3report,zai2026glm46v,internvl35report,minicpmv45report,gemma4docs}. The optimizer is fixed to Qwen3.5-9B, including the experiments in which it optimizes the larger Gemma4-31B worker. All methods compared within a model--task row use the same $400/200/1000$ train/validation/test split (SearchQA uses $1400$ test examples), a fixed split seed of $455$, and $12$ optimization steps. FinQA, CLUTRR, ChartQA, and ProofWriter are scored by accuracy; SearchQA by relaxed exact match. Local models are served with vLLM \citep{kwon2023efficient}, and the remote worker uses an OpenAI-compatible endpoint.

\subsection{Experimental Results}

\newcommand{\gain}[1]{$_{\scriptstyle\uparrow#1}$}
\newcommand{\loss}[1]{$_{\scriptstyle\downarrow#1}$}
\newcommand{\same}{$_{\scriptstyle\pm0.00}$}
\begin{table*}[!t]
\centering

\setlength{\tabcolsep}{4.6pt}
\small
\begin{tabular}{ll|ccc|lllll}
\Xhline{1.1pt}
\rowcolor{tabhead}
\textbf{Model} & \textbf{Method} & \textbf{Prompt} & \textbf{Skill} & \textbf{Route} & \textbf{FinQA} & \textbf{CLUTRR} & \textbf{ChartQA} & \textbf{ProofWriter} & \textbf{SearchQA} \\
\Xhline{1.1pt}
\multirow{7}{*}{\mlogo{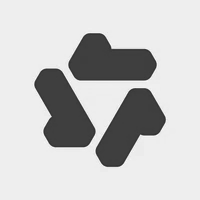}\,\textbf{Qwen3.5-4B}}
  & OPRO & \yes & \no & \no & 16.50 & 30.10 & 81.20 & 52.30 & 73.00 \\
  & TextGrad & \yes & \no & \no & 19.90\gain{3.40} & 31.10\gain{1.00} & 80.20\loss{1.00} & 73.20\gain{20.90} & 73.43\gain{0.43} \\
  & GEPA & \yes & \no & \no & 35.60\gain{19.10} & 28.00\loss{2.10} & 77.90\loss{3.30} & 54.00\gain{1.70} & 72.90\loss{0.10} \\
  & aPSF & \yes & \no & \no & 31.70\gain{15.20} & 32.30\gain{2.20} & 81.90\gain{0.70} & 60.90\gain{8.60} & 72.00\loss{1.00} \\
\cmidrule(lr){2-10}
  & Trace2Skill & \no & \yes & \no & 28.00\gain{11.50} & 65.20\gain{35.10} & 81.80\gain{0.60} & 71.40\gain{19.10} & 69.93\loss{3.07} \\
  & SkillOpt & \no & \yes & \no & 37.00\gain{20.50} & 62.40\gain{32.30} & 82.30\gain{1.10} & 75.40\gain{23.10} & 74.00\gain{1.00} \\
\cmidrule(lr){2-10}
\rowcolor{oursrow} & \textbf{\sparo{}} & \yes & \yes & \yes & \textbf{47.90}\gain{31.40} & \textbf{69.00}\gain{38.90} & \textbf{84.60}\gain{3.40} & \textbf{84.70}\gain{32.40} & \textbf{76.30}\gain{3.30} \\
\Xhline{0.8pt}
\multirow{7}{*}{\mlogo{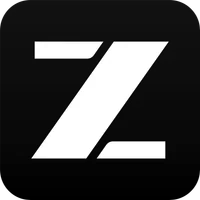}\,\textbf{GLM-4.6V-Flash}}
  & OPRO & \yes & \no & \no & 18.70 & 29.60 & 65.20 & 49.00 & 64.50 \\
  & TextGrad & \yes & \no & \no & 22.10\gain{3.40} & 29.30\loss{0.30} & 64.00\loss{1.20} & 50.50\gain{1.50} & 65.64\gain{1.14} \\
  & GEPA & \yes & \no & \no & 21.20\gain{2.50} & 32.00\gain{2.40} & 64.90\loss{0.30} & 31.20\loss{17.80} & 68.00\gain{3.50} \\
  & aPSF & \yes & \no & \no & 16.10\loss{2.60} & 29.00\loss{0.60} & 65.40\gain{0.20} & 47.80\loss{1.20} & 54.71\loss{9.79} \\
\cmidrule(lr){2-10}
  & Trace2Skill & \no & \yes & \no & 21.50\gain{2.80} & 28.60\loss{1.00} & 62.10\loss{3.10} & 43.50\loss{5.50} & 69.60\gain{5.10} \\
  & SkillOpt & \no & \yes & \no & 23.10\gain{4.40} & 28.00\loss{1.60} & 64.70\loss{0.50} & 59.20\gain{10.20} & 69.00\gain{4.50} \\
\cmidrule(lr){2-10}
\rowcolor{oursrow} & \textbf{\sparo{}} & \yes & \yes & \yes & \textbf{25.40}\gain{6.70} & \textbf{34.10}\gain{4.50} & \textbf{74.30}\gain{9.10} & \textbf{64.40}\gain{15.40} & \textbf{73.30}\gain{8.80} \\
\Xhline{0.8pt}
\multirow{7}{*}{\mlogo{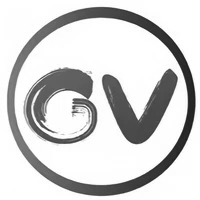}\,\textbf{InternVL3.5-8B}}
  & OPRO & \yes & \no & \no & 20.60 & 24.00 & 72.20 & 49.60 & 69.21 \\
  & TextGrad & \yes & \no & \no & 25.90\gain{5.30} & 25.70\gain{1.70} & 71.30\loss{0.90} & 44.50\loss{5.10} & 71.20\gain{1.99} \\
  & GEPA & \yes & \no & \no & 24.70\gain{4.10} & 25.70\gain{1.70} & 71.00\loss{1.20} & 30.90\loss{18.70} & 72.00\gain{2.79} \\
  & aPSF & \yes & \no & \no & 18.60\loss{2.00} & 25.90\gain{1.90} & 74.00\gain{1.80} & 41.30\loss{8.30} & 70.30\gain{1.09} \\
\cmidrule(lr){2-10}
  & Trace2Skill & \no & \yes & \no & 20.70\gain{0.10} & 25.80\gain{1.80} & 70.30\loss{1.90} & 37.90\loss{11.70} & 68.00\loss{1.21} \\
  & SkillOpt & \no & \yes & \no & 22.80\gain{2.20} & 26.20\gain{2.20} & 70.90\loss{1.30} & 47.40\loss{2.20} & 71.71\gain{2.50} \\
\cmidrule(lr){2-10}
\rowcolor{oursrow} & \textbf{\sparo{}} & \yes & \yes & \yes & \textbf{28.20}\gain{7.60} & \textbf{32.10}\gain{8.10} & \textbf{77.00}\gain{4.80} & \textbf{64.70}\gain{15.10} & \textbf{74.20}\gain{4.99} \\
\Xhline{0.8pt}
\multirow{7}{*}{\mlogo{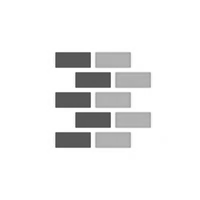}\,\textbf{MiniCPM-V-4.5}}
  & OPRO & \yes & \no & \no & 19.40 & 20.50 & 67.10 & 45.10 & 69.57 \\
  & TextGrad & \yes & \no & \no & 19.50\gain{0.10} & 20.10\loss{0.40} & 66.50\loss{0.60} & 52.70\gain{7.60} & 70.50\gain{0.93} \\
  & GEPA & \yes & \no & \no & 22.90\gain{3.50} & 22.70\gain{2.20} & 66.40\loss{0.70} & 38.10\loss{7.00} & 69.90\gain{0.33} \\
  & aPSF & \yes & \no & \no & 18.00\loss{1.40} & 19.80\loss{0.70} & 69.90\gain{2.80} & 48.80\gain{3.70} & 70.07\gain{0.50} \\
\cmidrule(lr){2-10}
  & Trace2Skill & \no & \yes & \no & 19.50\gain{0.10} & 18.30\loss{2.20} & 66.90\loss{0.20} & 59.00\gain{13.90} & 68.07\loss{1.50} \\
  & SkillOpt & \no & \yes & \no & 19.20\loss{0.20} & 24.30\gain{3.80} & 67.10\same & 44.50\loss{0.60} & 73.70\gain{4.13} \\
\cmidrule(lr){2-10}
\rowcolor{oursrow} & \textbf{\sparo{}} & \yes & \yes & \yes & \textbf{26.00}\gain{6.60} & \textbf{28.50}\gain{8.00} & \textbf{71.80}\gain{4.70} & \textbf{66.80}\gain{21.70} & \textbf{74.90}\gain{5.33} \\
\Xhline{0.8pt}
\multirow{7}{*}{\mlogo{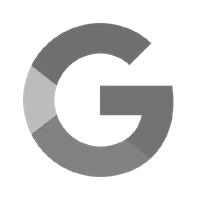}\,\textbf{Gemma4-31B}}
  & OPRO & \yes & \no & \no & 30.80 & 80.00 & 84.80 & 89.60 & 73.71 \\
  & TextGrad & \yes & \no & \no & 54.20\gain{23.40} & 90.40\gain{10.40} & 88.60\gain{3.80} & 92.80\gain{3.20} & 78.00\gain{4.29} \\
  & GEPA & \yes & \no & \no & 53.60\gain{22.80} & 92.00\gain{12.00} & 87.60\gain{2.80} & 88.80\loss{0.80} & 78.40\gain{4.69} \\
  & aPSF & \yes & \no & \no & 27.60\loss{3.20} & 85.80\gain{5.80} & 86.00\gain{1.20} & 91.80\gain{2.20} & 71.20\loss{2.51} \\
\cmidrule(lr){2-10}
  & Trace2Skill & \no & \yes & \no & 49.80\gain{19.00} & 90.40\gain{10.40} & 85.20\gain{0.40} & 92.20\gain{2.60} & 77.14\gain{3.43} \\
  & SkillOpt & \no & \yes & \no & 41.00\gain{10.20} & 86.00\gain{6.00} & 87.60\gain{2.80} & 93.60\gain{4.00} & 79.43\gain{5.72} \\
\cmidrule(lr){2-10}
\rowcolor{oursrow} & \textbf{\sparo{}} & \yes & \yes & \yes & \textbf{58.00}\gain{27.20} & \textbf{95.40}\gain{15.40} & \textbf{89.80}\gain{5.00} & \textbf{96.00}\gain{6.40} & \textbf{81.90}\gain{8.19} \\
\Xhline{1.1pt}
\end{tabular}
\caption{Main comparison across five worker models and five benchmarks.
The \textbf{Prompt}, \textbf{Skill}, and \textbf{Route} columns indicate which
components of the language program each optimizer can edit (\yes\ supported,
\no\ not supported). Within each model block, prompt-centered optimizers are
listed first in chronological order (OPRO, TextGrad, GEPA, aPSF), followed by
skill-centered optimizers (Trace2Skill, SkillOpt), then \sparo{}. OPRO is the
reference method in each block; subscripts give the absolute difference from it.
Best result per model--task setting in bold.}
\label{tab:main-results}
\end{table*}
Table~\ref{tab:main-results} reports the main comparison. \sparo{} obtains the highest point estimate in all 25 model--task settings, exceeding the strongest competing method by 4.4 points on average and by as much as 15.1 points. These are single-split-seed results; Appendix~G repeats the Qwen3.5-4B comparison against GEPA over three seeds, where \sparo{} leads on all five tasks and in all fifteen task--seed pairs. On the Qwen3.5-4B worker that margin averages 5.7 points across the five tasks.

\textbf{Joint optimization improves prompt-sensitive tasks over skill-only learning.} ChartQA and SearchQA place particular pressure on global task interpretation, evidence selection, and exact answer control. Across the ten worker-task pairs on these two benchmarks, \sparo{} improves over the stronger of Trace2Skill and SkillOpt by 3.7 points on average. The improvement is consistent in every pair, reaching 9.6 points for GLM-4.6V-Flash on ChartQA. These results indicate that reusable skills alone do not replace the need to optimize the global instruction and its interaction with routing.

\textbf{Joint optimization improves skill-intensive tasks over prompt-only search.} FinQA, CLUTRR, and ProofWriter contain recurring arithmetic, relational, and deductive operations that can be represented as reusable procedures. Across all 15 worker-task pairs on these benchmarks, \sparo{} exceeds the strongest prompt-centered baseline among GEPA, aPSF, OPRO, and TextGrad by 9.1 points on average. The contrast is largest for Qwen3.5-4B on CLUTRR, where \sparo{} reaches 69.0 while the strongest prompt-centered method reaches 32.3. The result supports the complementary claim: global prompt optimization is insufficient when recurring failures can be localized into reusable skills.

\textbf{A small optimizer can guide a substantially larger worker.} With Qwen3.5-9B fixed as the optimizer, \sparo{} obtains the best result for the Gemma4-31B worker on all five benchmarks. It improves over the strongest competing method by 2.7 points on average. Thus, \sparo{}'s benefit does not require the optimizer to be larger than the worker; structured attribution can allow a smaller optimizer to provide useful program-level supervision to a stronger model.

\subsection{Component Selection Patterns Across Tasks}

We record the component selected at each optimization step and compute
how often each of prompt, skill, and routing is chosen.

\begin{figure}[!htbp]
    \centering
    \includegraphics[width=\columnwidth]{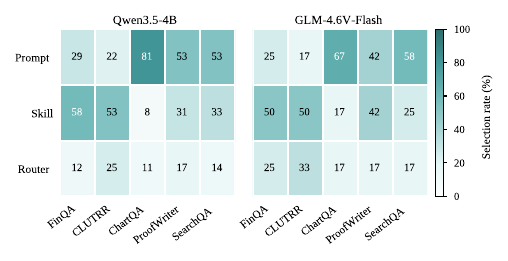}
    \caption{
    Component-selection patterns across tasks for the Qwen3.5-4B (left)
    and GLM-4.6V-Flash (right) workers. Each column gives the percentage of
    optimization steps that selected a prompt, skill, or routing mutation,
    so the three cells of a column sum to $100\%$ up to rounding.
    }
    \label{fig:component_selection_heatmap}
\end{figure}

Figure~\ref{fig:component_selection_heatmap} shows that \sparo{} adapts
its component choices to both the task and the worker. Skill mutations
are selected most often on CLUTRR and FinQA for both workers,
consistent with recurring relational and numerical reasoning
patterns. Prompt mutations dominate on ChartQA and lead on ProofWriter
and SearchQA for the Qwen worker, where global task interpretation and
answer construction matter more. Routing mutations are never the most
frequent choice, but they keep a non-trivial share in every setting,
from $11.1\%$ to $33.3\%$, so the selector spreads its edits over all
three components rather than collapsing onto one.

\section{Ablation Study}

\subsection{System-Level Component Ablation}

Table~\ref{tab:system-ablation} reports the performance degradation
relative to the complete \sparo{} configuration after removing each major
component, using the Qwen3.5-4B worker. Full \sparo{} obtains accuracies of
$69.0$ on CLUTRR and $84.7$ on ProofWriter.

\begin{table}[!htbp]
\centering
\small
\setlength{\tabcolsep}{8pt}
\begin{tabular}{lrr}
\toprule
 & CLUTRR & ProofWriter \\
\midrule
 w/o Attribution
& 28.2
& 83.1 \\

 w/o Skill
& 31.2
& 77.1 \\

 w/o Cross-Round Memory
& 62.9
& 80.7 \\
 w/o Routing
& 49.7
& 75.5 \\

\textbf{\sparo{} (Ours)}
& \textbf{69.0}
& \textbf{84.7} \\
\bottomrule
\end{tabular}
\caption{System-level ablation results on CLUTRR and ProofWriter. Values are accuracies in percentages.}
\label{tab:system-ablation}
\end{table}
Across the two benchmarks, removing skills and removing counterfactual
attribution are the two most damaging changes, costing $22.7$ and $21.2$
points on average: the optimizer needs both a place to store reusable
procedural knowledge and a way to decide which component to edit.
Removing routing costs $14.3$ points, indicating that stored skills must
also be activated on the right inputs, while the cross-round memory adds
a smaller but consistent $5.1$ points. Every component contributes
positively.

\subsection{Attribution Strategy Analysis}
\begin{figure}[!htbp]
    \centering
    \includegraphics[width=0.74\columnwidth]{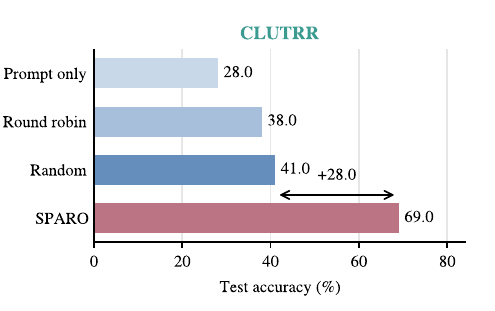}
    \caption{
    Comparison of component-selection strategies on CLUTRR.
    \sparo{} outperforms random selection, round-robin selection, and
    prompt-only optimization.
    }
    \label{fig:finqa_attribution_strategy}
\end{figure}
We compare \sparo{}'s counterfactual attribution mechanism with three
simpler component-selection strategies on CLUTRR with the Qwen3.5-4B
worker: random selection, round-robin selection, and prompt-only
optimization. All variants share the modular representation, the three
mutation operators, the optimizer model, the data split, the step budget
and the candidate-pool selection procedure, and differ only in how the
component to edit is chosen. As shown in
Figure~\ref{fig:finqa_attribution_strategy}, \sparo{} reaches $69.0$ against
$41.0$, $38.0$, and $28.0$, so effective modular optimization depends on
using counterfactual evidence to identify the component to update rather
than on random, fixed-order, or prompt-exclusive selection.

\section{Parameter Sensitivity}

\begin{figure}[t]
    \centering
    \includegraphics[width=\columnwidth]{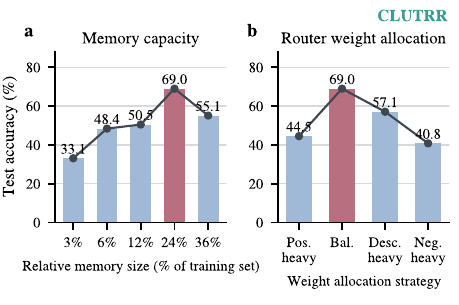}
    \caption{
    Parameter sensitivity of \sparo{} on CLUTRR. We vary (a) memory capacity
    relative to the training-set size and (b) router weight allocation:
    balanced, and positive-, description- and negative-heavy.
    }
    \label{fig:parameter_sensitivity}
\end{figure}

We analyze the sensitivity of \sparo{} to two hyperparameters associated
with its main components: the capacity of the cross-round memory and the router weight allocation. All runs use the Qwen3.5-4B worker on CLUTRR, with every other hyperparameter held fixed. The number of experience clusters and the attribution temperature are analyzed in Appendix E.

Figure~\ref{fig:parameter_sensitivity} shows that both parameters have an
interior optimum. Small memory capacities provide insufficient historical
evidence, while an excessively large memory may introduce redundant or
outdated experiences; a capacity of $24\%$ of the training set performs
best.

The balanced configuration $(w_+,w_d,w_-)=(0.55,0.25,0.40)$ scores highest
at $69.0$, ahead of description-heavy $(0.40,0.45,0.30)$ at $57.1$,
positive-heavy $(0.70,0.20,0.30)$ at $44.5$, and negative-heavy
$(0.50,0.20,0.60)$ at $40.8$. Tilting toward any one signal costs accuracy,
though the sweep rebalances rather than removes a signal and so does not
isolate each term.

\section{Conclusion}

We presented \sparo{}, which treats the prompt, the skill library, and the
routing rules of an LLM program as jointly optimized components. Using a
cross-round experience pool and counterfactual attribution, \sparo{} decides
where a failure originates before deciding how to revise the program, and
improves over recent prompt- and skill-oriented optimizers across five
benchmarks and five worker models.

\bibliography{aaai2027}

\end{document}


\maketitle
\enlargethispage{-46pt}

\appendix

\section{A: Detailed Algorithms}
\label{app:detailed-algorithms}

This section provides the complete algorithmic details of the SPARO
Optimization Loop (see Algorithm 1).

\begin{algorithm}[!t]
\caption{SPARO Optimization Loop}
\label{alg:sparo}
\begin{algorithmic}[1]
\Require Initial program
$P_0=(p_0,\mathcal B_0,r_0,t_0)$,
training and validation sets, optimization steps $K$
\Ensure Optimized program $P^\star$

\State Evaluate $P_0$ on the validation set
\State Initialize candidate pool $\mathcal C\gets\{P_0\}$,
Pareto frontier $\mathcal H\gets\{P_0\}$, and
experience memory $\mathcal M$

\For{$k=1,\ldots,K$}
    \State Select parent program $P_k\in\mathcal C$ using
    validation quality and diversity
    \State Sample training minibatch $\mathcal D_k$ and execute
    $P_k$ to collect predictions, scores, and traces
    \State Insert new experiences into $\mathcal M$ using
    cluster-balanced replacement
    \State Re-evaluate $\mathcal M$ under $P_k$ and construct
    failure and success sets $\mathcal F_k$ and $\mathcal S_k$

    \State Retrieve 
    evidence $\mathcal E_k$
    \State Evaluate prompt-repair, routing-add/drop, and
    skill-repair/new-skill counterfactuals on $\mathcal E_k$
    \State Compute
    \[
    \bar P_k(c;\tau)
    \gets
    \frac{1}{|\mathcal E_k|}
    \sum_{e\in\mathcal E_k}P(c\mid e;\tau)
    \]
    \State Sample
    \[
c_k^\ast
\sim
\mathrm{Categorical}
\left(
\left\{
\bar{P}_k(c;\tau)
\right\}_{c\in C}
\right)
    \]

    \State Collect component-specific mutation evidence
    $\mathcal E_{c_k^*}$ from the retrieved clusters
    \State Generate
    \[
    \widetilde P_k
    \gets
    \textsc{Mutate}
    (P_k,c_k^*,\mathcal E_{c_k^*})
    \]
    \State Evaluate $\widetilde P_k$ on the validation set
    \State Update candidate pool $\mathcal C$ and Pareto
    frontier $\mathcal H$ using validation-based Pareto
    \State Record attribution, mutation, and evaluation metadata
\EndFor
 \Return
\[
P^\star
=
\arg\max_{P\in\mathcal C}J_{\mathrm{val}}(P)
\]
\end{algorithmic}
\end{algorithm}

\section{B: Reproducibility Checklist}
\label{app:reproducibility}

We provide comprehensive experimental details to ensure full
reproducibility of our results:

\begin{itemize}
    \item \textbf{Datasets:} All benchmark names, task types, data
    sources, split construction, evaluation metrics, and license notes
    are provided in Appendix C. We use the same
    training, validation, and test partitions for all methods within
    each model--task setting. FinQA, CLUTRR,
    ChartQA, and ProofWriter use 400 training examples, 200 validation
    examples, and 1000 test examples; SearchQA uses 400 training
    examples, 200 validation examples, and 1400 test examples.

    \item \textbf{Model Configurations:} Complete model configurations
    are fixed within each model--task row. Worker models include
    Qwen3.5-4B, GLM-4.6V-Flash, InternVL3.5-8B, MiniCPM-V-4.5, and
    Gemma4-31B. The optimizer LLM is Qwen3.5-9B in all experiments,
    including the settings where it optimizes the larger Gemma4-31B
    worker. Local models are served with vLLM through OpenAI-compatible
    endpoints, while Gemma4-31B is accessed through a remote
    OpenAI-compatible API. Worker inference is deterministic with
    temperature 0.0; optimizer calls use temperature 0.7.

    \item \textbf{Evaluation Protocol:} All methods compared within a
    row use identical worker models, optimizer models, data splits,
    decoding settings, and 12 optimization steps. Candidate programs are
    selected by validation performance, and the held-out test set is
    evaluated only for the seed program and the final
    validation-selected program. FinQA, CLUTRR, ChartQA, and ProofWriter
    use accuracy; SearchQA uses relaxed exact match to account for
    answer-span normalization.

    \item \textbf{Prompts and Meta-Prompts:} The release contains the
    initial task prompts and component-specific mutation prompts. Sample meta-prompts are provided in Appendix F.

    \item \textbf{Hyperparameters:} SPARO uses 12 optimization steps, 
32 examples as a minibatch in each step, and at most 8 active skill
blocks during optimization. SPARO maintains a
cluster-balanced experience memory, where the memory capacity is set to
24\% of the training set size according to sensitivity analysis. With
this capacity fixed, the number of experience clusters is set to 5,
which provides the best trade-off between evidence diversity and
within-cluster support. The router adopts the balanced configuration
$(w_+,w_d,w_-)=(0.55,0.25,0.40)$ for positive triggers,
semantic descriptions, and negative triggers. The attribution
temperature is set to $\tau=0.5$, controlling the concentration of
component responsibility during cross-example attribution aggregation. The detailed hyperparameter sensitivity study of number of clusters and attribution temperature is on appendix E.

    \item \textbf{Random Seeds:} We use a fixed split seed
    \texttt{455}. For datasets with official train/validation/test
    splits, examples are shuffled deterministically and sampled with
    seeds 455, 456, and 457 for train, validation, and test,
    respectively. For CLUTRR, the official training file supplies the
    training subset, and the official test file is shuffled once and
    partitioned into validation and test subsets. Worker decoding is
    deterministic; optimizer stochasticity is controlled by the run seed
    and recorded in each run command. We also conduct a three-seed comparison between SPARO and GEPA using
seeds 79, 123, and 455. Details are provided in Appendix G.

    \item \textbf{Computational Resources:} All experiments run on the following infrastructure:

    \begin{itemize}
    \item \textbf{GPUs}: 2 NVIDIA A100 80GB   GPUs for worker inference
    \item \textbf{CPU}: 64-core AMD EPYC 7763 @ 2.45GHz
    \item \textbf{RAM}: 512GB DDR4
    \item \textbf{OS}: Ubuntu 20.04 LTS, CUDA 11.8, PyTorch 2.0.1
    \end{itemize}

    \item \textbf{Code and Artifacts Release:} Upon paper acceptance, we will release: (1) complete source code with documentation, (2) all meta-prompts as .txt files, (3) validation/test indices as JSON, (4) discovered factor structures per benchmark, (5) optimization trajectories (per-iteration metrics), and (6) bash scripts for reproducing all tables/figures. Repository will be hosted on GitHub under MIT license.
\end{itemize}
\begin{table*}[t]
\centering
\small
\setlength{\tabcolsep}{4pt}

\begin{tabular}{
>{\raggedright\arraybackslash}p{0.12\linewidth}
>{\raggedright\arraybackslash}p{0.19\linewidth}
>{\raggedright\arraybackslash}p{0.21\linewidth}
>{\raggedright\arraybackslash}p{0.10\linewidth}
>{\raggedright\arraybackslash}p{0.28\linewidth}
}

\toprule
Dataset 
& Task Type 
& Source and Split 
& Metric 
& Reproducibility Notes \\
\midrule

FinQA
& Financial numerical reasoning over reports and tables
& Official train/dev/test files
& Accuracy
& We sample 400/200/1000 train/validation/test examples for local workers. \\

CLUTRR
& Relational reasoning and rule induction
& Task ID 1.5; official train file and official test file
& Accuracy
& Training examples are sampled from the official train file. The official test file is deterministically shuffled and partitioned into validation and held-out test subsets. \\

ChartQA
& Visual chart understanding and answer formatting
& Human subset with official train/validation/test directories
& Accuracy
& We use the human subset and keep image-question-answer triples aligned with their original chart files. All methods use identical sampled examples. \\

ProofWriter
& Multi-step deductive reasoning over facts and rules
& Public JSONL splits
& Accuracy
& Inputs are converted into a unified text format containing facts, rules, and the target query. Splits are sampled deterministically with the shared seed. \\

SearchQA
& Noisy search-snippet question answering
& SearchQA item splits materialized from fixed ID files
& Relaxed exact match
& We use 400/200/1400 train/validation/test examples. Evaluation normalizes extracted spans and accepts equivalent short answer strings. \\

\bottomrule

\end{tabular}

\caption{
Dataset sources and split construction. All benchmarks are publicly available; licenses and redistribution terms follow the original dataset releases. The code release will include download instructions, preprocessing scripts, and the exact sampled indices used in all experiments.
}

\label{tab:dataset-sources}

\end{table*}
\section{C: Datasets and Splits}
\label{app:datasets}

We evaluate on five public benchmarks that cover complementary
reasoning and question-answering regimes. Table~\ref{tab:dataset-sources}
summarizes the task type, source split, evaluation metric, and
reproducibility notes for each benchmark. No new dataset is introduced
in this work.

All data preprocessing is deterministic. For text-only benchmarks, the
preprocessing scripts convert each record into a common example object
containing the input text, gold answer, and metadata needed by the
evaluator. For ChartQA, the scripts preserve the image path and chart
question while applying the same sampling protocol as the text
benchmarks. For SearchQA, fixed ID splits are materialized into
train/validation/test item files before optimization so that all methods
read the same examples. 

\section{D: Model Configurations}
\label{app:model-config}

This section follows the role-based reporting style used in prior
prompt-optimization appendices: we separate the LLM that executes the
task from the LLM that proposes optimization edits. This distinction is
important for SPARO because all compared methods in a model--task row
share the same worker model, optimizer model, API interface, and decoding
settings; only the optimization algorithm differs.

\subsection{D.1 LLM Roles and Specifications}

SPARO uses two LLM roles. The worker model $M_W$ executes the assembled
program on benchmark examples. The optimizer model $M_O$ reads traces,
failure evidence, and component metadata, then proposes prompt, skill,
or routing edits. Table~\ref{tab:model-specs} summarizes the model
identifiers and deployment settings used in the main experiments.

\begin{table}[!htbp]
\centering
\small
\setlength{\tabcolsep}{4pt}

\begin{tabular}{
>{\raggedright\arraybackslash}m{0.28\columnwidth}
>{\raggedright\arraybackslash}m{0.64\columnwidth}}
\toprule
Role & Model configuration \\
\midrule

Default worker
&
Qwen3.5-4B; text and image;
\texttt{Qwen3\_5For\allowbreak ConditionalGeneration};
262K-token context; served through local vLLM. \\
\cmidrule(lr){1-2}

Optimizer
&
Qwen3.5-9B; text optimizer;
\texttt{Qwen3\_5For\allowbreak ConditionalGeneration};
262K-token context; served through local vLLM and used for SPARO and
all baseline optimization calls. \\
\midrule

\multicolumn{2}{l}{\textit{Alternative workers}} \\
\midrule

InternVL3.5-8B
&
Vision--language model; \texttt{InternVLChatModel};
bfloat16 weights; 40K-token context; served through local vLLM. \\
\cmidrule(lr){1-2}

MiniCPM-V-4.5
&
Vision--language model; \texttt{MiniCPMV};
bfloat16 weights; 40K-token context; served through local vLLM. \\
\cmidrule(lr){1-2}

GLM-4.6V
&
Vision--language model;
\texttt{Glm4v\allowbreak ForConditionalGeneration};
131K-token context; served through local vLLM. \\
\cmidrule(lr){1-2}

Gemma4-31B
&
Text-and-image model accessed through a remote OpenAI-compatible API;
provider-side precision and context settings are not locally
inspectable. \\

\bottomrule
\end{tabular}

\caption{Model roles and configurations used in the experiments.}
\label{tab:model-specs}
\end{table}

\subsection{D.2 Serving Interface}

All local models are served through OpenAI-compatible vLLM endpoints.
For the default Qwen3.5 experiments, the worker endpoint is
\texttt{http://127.0.0.1:8015/v1} and the optimizer endpoint is
\texttt{http://127.0.0.1:8004/v1}. Alternative local workers reuse the
same task endpoint role while keeping the optimizer endpoint fixed to
Qwen3.5-9B. For remote Gemma4-31B experiments, the worker endpoint is
replaced by the remote OpenAI-compatible API, while optimization calls
continue to use the local Qwen3.5-9B endpoint. This protocol isolates
the effect of changing the worker model from the effect of changing the
optimizer.

\subsection{D.3 Decoding Hyperparameters}

Table~\ref{tab:decoding-config} reports the decoding settings shared by
SPARO and all baselines. Worker inference is deterministic so that
validation comparisons reflect candidate quality rather than sampling
noise. Optimizer calls use a nonzero temperature to generate diverse
candidate edits.

No explicit chain-of-thought or thinking-mode token is inserted into the
worker prompts. The worker is instructed to return the final answer in a
fixed answer tag, while the optimizer receives meta-prompts for program
editing. For baseline fairness, OPRO, GEPA, TextGrad, aPSF, Trace2Skill,
SkillOpt, and SPARO use the same worker and optimizer decoding settings
when evaluated under the same model row.

\begin{center}
\small
\setlength{\tabcolsep}{4pt}

\begin{tabular}{
>{\raggedright\arraybackslash}p{0.34\columnwidth}
>{\raggedright\arraybackslash}p{0.26\columnwidth}
>{\raggedright\arraybackslash}p{0.27\columnwidth}}
\toprule
Parameter & Worker $M_W$ & Optimizer $M_O$ \\
\midrule
Temperature
& 0.0
& 0.7 \\

Maximum output tokens
& 8192
& 8192 \\

Top-$p$ / top-$k$
& Server default
& Server default \\

Repetition penalty
& Server default
& Server default \\

Prompt budget
& Bounded by task endpoint context
& Bounded by optimizer endpoint context \\
\bottomrule
\end{tabular}

\captionof{table}{Decoding hyperparameters.}
\label{tab:decoding-config}
\end{center}

\section{E: Hyperparameter Sensitivity Study}
\begin{figure}[!htbp]
    \centering
    \includegraphics[width=\columnwidth]{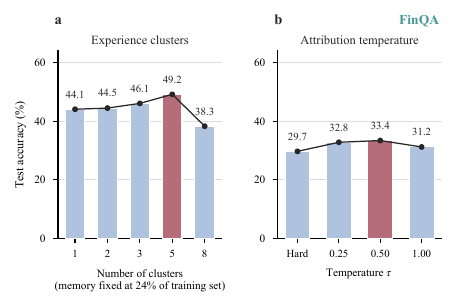}
    \caption{
    Parameter sensitivity of SPARO on FinQA, with the memory capacity held
    at its selected size. We vary (a) the number of experience clusters and
    (b) the attribution temperature; ``Hard'' denotes $\tau\to0$.
    }
    \label{fig:parameter_sensitivity}
\end{figure}
We analyze the sensitivity of SPARO to two additional hyperparameters:
the number of experience clusters and the attribution temperature.
All experiments are conducted on the same dataset while keeping the
worker model, optimizer model, data split, optimization budget, and
remaining hyperparameters fixed.

With the memory capacity fixed at the best size, performance
improves as the number of clusters increases from one
to five. Five clusters yield the best result, corresponding to
an average per-cluster capacity equal to $4.8\%$ of the
training-set size. Increasing the number of clusters to eight reduces
the average per-cluster capacity to $3\%$ and substantially degrades
performance, indicating that overly fine partitioning
can leave insufficient evidence within each cluster.

Attribution temperature also exhibits an intermediate optimum which
performs better than both hard attribution and the
smoother configuration. Overall, these results suggest that
SPARO is most effective when its memory, clustering granularity,
routing signals, and attribution uncertainty are moderately
balanced.
\section{F: Meta-Prompts}
\label{app:meta-prompts}

This section reports the normalized meta-prompt templates used by SPARO.
The templates describe the instruction structures used by the optimizer
LLM for assembling programs and generating component-specific mutations.
The released artifact will include the exact implementation prompts.

\subsection{F.1 Program Assembly Template}
\label{app:meta-program-assembly}

SPARO separates optimization metadata from execution. The worker LLM
receives only the assembled program consisting of the global prompt and
the subset of skill blocks selected by the router. It does not receive
attribution scores, failure clusters, or mutation information.

\begin{quote}
\small
\textbf{System prompt template.}

\[
\begin{array}{l}
\texttt{\{main\_prompt\}}\\
\texttt{[optional: selected skill blocks]}
\end{array}
\]

When skill blocks are activated, SPARO appends the selected procedural
knowledge as auxiliary guidance. Each skill block contains its name,
description, activation conditions, and execution procedure. The main
prompt remains responsible for global task instructions, while skill
blocks provide local reusable knowledge for specific input patterns.
\end{quote}

\subsection{F.2 Prompt Mutation Meta-Prompt}
\label{app:meta-prompt-mutation}

Prompt mutation is invoked when counterfactual attribution indicates
that the global prompt is the most responsible component. The optimizer
LLM modifies only the main prompt while preserving the existing skill
library and routing mechanism.

\begin{quote}
\small

\textbf{Role.}
You are the Prompt Mutation module of SPARO. Your goal is to improve the
global task instruction according to the identified failure patterns.

\textbf{Inputs.}
The prompt receives the current main prompt, the selected component
attribution result, component-specific evidence examples retrieved from
the experience memory, and validation examples associated with the
selected failure clusters.

\textbf{Editing rules.}
The global prompt should encode task-level requirements, general
reasoning strategies, and output constraints. The optimizer may refine
ambiguous instructions or introduce missing general rules. It should not
memorize individual answers, copy failure examples, or move reusable
procedures into the global prompt.

\textbf{Output constraint.}
Return only the revised main prompt text without additional explanation.

\end{quote}

\subsection{F.3 Skill Mutation Meta-Prompt}
\label{app:meta-skill-mutation}

Skill mutation is invoked when attribution indicates that the skill
library should be modified. SPARO supports both repairing existing skills
and proposing new reusable skills when the current library does not
contain sufficient knowledge.

\begin{quote}
\small

\textbf{Role.}
You are the Skill Mutation module of SPARO. Your goal is to maintain a
compact library of reusable procedural knowledge.

\textbf{Inputs.}
The prompt receives the current skill library, the selected component
attribution result, retrieved failure clusters, and successful examples
used as regression references.

\textbf{Skill design rules.}
Each skill should describe one reusable and verifiable behavior. The
optimizer may revise an existing skill when its procedure is incorrect
or incomplete, or create a new skill when the required knowledge is not
covered by the current library. Skills should contain a concise
description, activation conditions, and an execution procedure. They
should not memorize exact answers, evaluator feedback, or previous model
outputs.

\textbf{Output constraint.}
Return only the updated skill block(s) in the predefined skill format.

\end{quote}

\subsection{F.4 Routing Mutation Meta-Prompt}
\label{app:meta-routing-mutation}

Routing mutation is invoked when attribution indicates that the skill
content is useful but the activation policy is incorrect. The optimizer
updates routing conditions while keeping the underlying skill content
fixed.

\begin{quote}
\small

\textbf{Role.}
You are the Routing Mutation module of SPARO. Your goal is to improve
when reusable skills are activated.

\textbf{Inputs.}
The prompt receives the target skill, its description and procedure,
current routing conditions, positive examples where the skill should be
activated, and negative examples where activation is undesirable.

\textbf{Routing rules.}
The optimizer should identify observable input patterns that distinguish
successful and unsuccessful activation cases. It may add missing
activation conditions or remove overly broad conditions that cause
harmful skill usage. Routing conditions should depend on input
characteristics rather than evaluator labels or hidden optimization
metadata.

\textbf{Output constraint.}
Return only the revised routing conditions without explanation.

\end{quote}

\section{G: Three-Seed SPARO--GEPA Comparison}
\label{app:qwen35-three-seed}

This section reports the three-seed comparison between SPARO and GEPA
under the Qwen3.5-4B worker setting. All entries are test-set scores on a
0--100 scale, where FinQA, CLUTRR, ChartQA, and ProofWriter use
accuracy, and SearchQA uses relaxed exact match. We do the paired
SPARO/GEPA experiments for seeds 79, 123, and 455
under the same data sizes and optimization budget used in the main
experiments.

\begin{table*}[!t]
\centering
\small
\setlength{\tabcolsep}{3.2pt}

\begin{tabular}{lrrrrrrrrr}
\toprule
& \multicolumn{4}{c}{SPARO}
& \multicolumn{4}{c}{GEPA}
& \\
\cmidrule(lr){2-5}
\cmidrule(lr){6-9}
Task
& Seed 79 & Seed 123 & Seed 455 & Avg.
& Seed 79 & Seed 123 & Seed 455 & Avg.
& Gain \\
\midrule

FinQA
& 37.40 & 33.60 & 47.90 & 39.63
& 23.00 & 26.20 & 35.60 & 28.27
& +11.37 \\

CLUTRR
& 64.90 & 60.50 & 69.00 & 64.80
& 33.80 & 36.20 & 28.00 & 32.67
& +32.13 \\

ChartQA
& 86.60 & 89.80 & 84.60 & 87.00
& 78.50 & 79.90 & 77.90 & 78.77
& +8.23 \\

ProofWriter
& 70.40 & 74.70 & 84.70 & 76.60
& 55.70 & 68.40 & 54.00 & 59.37
& +17.23 \\

SearchQA
& 75.36 & 75.87 & 76.30 & 75.84
& 73.00 & 75.50 & 72.90 & 73.80
& +2.04 \\

\midrule

Macro Avg.
& 66.93 & 66.89 & 72.50 & 68.77
& 52.80 & 57.24 & 53.68 & 54.57
& +14.20 \\

\bottomrule
\end{tabular}

\caption{Three-seed test performance of SPARO and GEPA with the Qwen3.5
worker model. ``Gain'' denotes the difference between the SPARO and GEPA
three-seed averages, measured in percentage points.}
\label{tab:qwen35-three-seed-sparo-gepa}
\end{table*}

SPARO outperforms GEPA on every task and every reported seed. The
largest average gains appear on CLUTRR (+32.13 points), ProofWriter
(+17.23 points), and FinQA (+11.37 points), where failures often require
reusable procedural knowledge or corrected component usage rather than a
single global prompt rewrite. The gain is smaller on ChartQA
(+8.23 points) and SearchQA (+2.04 points), suggesting that GEPA already
captures much of the prompt-level behavior on these tasks, while SPARO
still provides a consistent benefit by combining prompt, skill, and
routing updates. Across the five tasks, SPARO improves the macro-average
test score from 54.57 to 68.77, corresponding to a 14.20-point gain.

\section{H: Qualitative Case Studies}
\label{app:case-studies}

This section presents three qualitative examples from SPARO optimization
logs. Each case follows the same sequence: an observed failure, the
evidence available to SPARO, the component-selection scores, the selected
mutation, the concrete program edit, and the resulting behavioral
change.

The tables below report the component-selection probabilities
$\bar{P}_k(c;\tau)$ defined in the main text. They are non-negative and
sum to one. ``Selected'' is the component sampled in that run; under
categorical sampling it need not be the highest-probability component,
although it is in each of the three cases shown.

\subsection{Prompt Attribution Case}
\label{app:prompt-case}

\paragraph{Initial failure.}
In a SearchQA run with Qwen3.5-4B as the worker model, the seed program
returned an overly long document title instead of the entity requested
by the clue.

\begin{quote}
\small
\textbf{Example Input.}
SearchQA trace excerpt:
\texttt{Things Overheard During Dick Cheney's Hunting Trip
... ``Has everyone updated their will?'' ...}\\
\textbf{Expected answer.} \texttt{Dick Cheney}\\
\textbf{Initial prediction.}
\texttt{Top Ten Things Overheard During Dick Cheney's Hunting Trip}
\end{quote}

\paragraph{Retrieved evidence.}
For the component decision in this run, SPARO's selector recorded
48 failure examples and 48 success anchors. The available aggregate diagnostics show
that the prompt intervention repaired 12 selected failures while
regressing on 3 selected successes.

\paragraph{Attribution.}
The Prompt component receives the highest 
component-selection score and is selected.

\begin{table}[t]

\centering
\small
\setlength{\tabcolsep}{4pt}
\begin{tabular}{lcccl}
\toprule
Case & Prompt & Skill & Routing & Selected \\
\midrule
SearchQA-123
& \textbf{0.552}
&  0.316
&  0.132
& \textsc{PromptEdit} \\
\bottomrule
\end{tabular}
\caption{Prompt-case component-selection scores after controlled
counterfactual checks.}
\label{tab:app-prompt-attribution-case}
\end{table}

\paragraph{Diagnosis.}
The failure reflects a global answer-contract error: the worker
identifies the relevant topic but returns a descriptive document title
rather than the shortest canonical answer span. The Skill component also
receives a positive score, but its repair count is smaller, while the
Routing component provides no repair evidence for this decision. SPARO
therefore treats the error as a weak global extraction policy rather
than missing reusable knowledge or a missed skill activation.

\paragraph{Mutation.}
SPARO selects \textsc{PromptEdit} and rewrites the global prompt to make
answer granularity explicit.

\begin{quote}
\small
\textbf{Before.}
\texttt{Find the shortest answer span or entity that answers the
question. Return only the final answer inside
\textless answer\textgreater...\textless/answer\textgreater.}\\
\textbf{After.}
\texttt{Extract the shortest, most canonical answer span that directly
answers the question. Select the specific entity or phrase rather than
a longer descriptive title. Output strictly
\textless answer\textgreater[CONTENT]\textless/answer\textgreater.}
\end{quote}

\paragraph{Effect.}
The final selected program corrects the example above with the
prediction \texttt{Dick Cheney}. At the run level, validation relaxed
exact match improves from 0.705 to 0.730, while held-out test relaxed
exact match improves from 0.735 to 0.763.

\subsection{Skill Attribution Case}
\label{app:skill-case}

\paragraph{Initial failure.}
In a FinQA run, the seed program failed on a reusable numerical pattern:
computing a range from financial table values while respecting the
stated unit scale.

\begin{quote}
\small
\textbf{Example Input.}
FinQA trace excerpt:
\texttt{chemicals of 2016 is 3474; chemicals of 2015 is 3543;
chemicals of 2014 is 3664. Question: what is the mathematical range for
chemical revenue from 2014--2016, in millions?}\\
\textbf{Expected answer.} \texttt{190}\\
\textbf{Initial prediction.} \texttt{1190}\\

\end{quote}

\paragraph{Retrieved evidence.}
For the final component decision records 45 selected failures and 51 success
anchors. The learned skill proposal
\texttt{financial\_unit\_scale\_validator} was created from SPARO's
missing-skill attribution. The failure rescue therefore supports a reusable-pattern diagnosis.

\paragraph{Attribution.}
Skill component receives the highest score
and is selected.

\begin{table}[t]

\centering
\small
\setlength{\tabcolsep}{4pt}
\begin{tabular}{lcccl}
\toprule
Case & Prompt & Skill & Routing & Selected \\
\midrule
FinQA-455
& 0.324
& \textbf{0.676}
& 0
& \textsc{SkillEdit} \\
\bottomrule
\end{tabular}
\caption{Skill-case component-selection scores after controlled
counterfactual checks.}
\label{tab:app-skill-attribution-case}
\end{table}

\paragraph{Diagnosis.}
The error is better represented as reusable procedural knowledge than
as a one-off prompt rewrite. FinQA repeatedly requires the worker to
check table units, identify the requested output scale, and perform
arithmetic under that scale. Prompt repair fixes some failures but also
regresses on 8 success anchors, while the Routing intervention repairs
no failures and regresses on 1 success anchor. SPARO therefore selects
\textsc{SkillEdit}.

\paragraph{Mutation.}
SPARO creates a new reusable skill rather than extending the global
prompt.

\begin{quote}
\small
\textbf{Name.}
\texttt{financial\_unit\_scale\_validator}\\
\textbf{Description.}
Enforces correct magnitude extraction and unit conversion for financial
table data.\\
\textbf{When to use.}
Triggered by financial tables with explicit units such as
\texttt{\$ in millions}, \texttt{in millions},
\texttt{financial table}, or \texttt{total of}.\\
\textbf{Procedure.}
Scan table headers and rows for unit declarations, extract the target
metric, determine whether the question requests raw or scaled units,
apply the corresponding conversion, and return only the final answer.\\
\textbf{Verification boundary.}
Check that the final magnitude is consistent with the unit declared in
the source table.
\end{quote}

\paragraph{Effect.}
The final selected program corrects the example above with the
prediction \texttt{190}. The iteration that first introduces the
unit-scale skill improves validation accuracy from 0.180 to 0.205. The
final validation-selected program improves validation accuracy from
0.180 to 0.480 and held-out test accuracy from 0.166 to 0.479.

\subsection{Routing Attribution Case}
\label{app:routing-case}

\paragraph{Initial failure.}
A SearchQA routing failure occurs when useful extraction knowledge
already exists in the skill library but the activation policy does not
reliably expose it to the worker.

\begin{quote}
\small
\textbf{Example Input.}
SearchQA trace excerpt:
\texttt{Goodyear Tire and Rubber Company ... On November
20, 1986, Goodyear acquired ...}\\
\textbf{Expected answer.} \texttt{Goodyear}\\
\textbf{Initial prediction.}
\texttt{Goodyear Tire and Rubber Company}\\
\textbf{Relevant existing skill.}
\texttt{concise\_entity\_extraction}
\end{quote}

\paragraph{Counterfactual evidence.}
After injecting the relevant existing skill
\texttt{concise\_entity\_extraction}, the available diagnostics show
that the routing-add intervention repaired 10 selected failures while
regressing on 2 selected successes.

\paragraph{Attribution.}
The Routing component receives the highest 
component-selection score and is selected.

\begin{table}[t]

\centering
\small
\setlength{\tabcolsep}{4pt}
\begin{tabular}{lcccl}
\toprule
Case & Prompt & Skill & Routing & Selected \\
\midrule
SearchQA-455
&  0.351
&  0.232
& \textbf{0.417}
& \textsc{RoutingEdit} \\
\bottomrule
\end{tabular}
\caption{Routing-case component-selection scores after controlled
counterfactual checks.}
\label{tab:app-routing-attribution-case}
\end{table}

\paragraph{Diagnosis.}
The relevant procedural knowledge is already present in the skill
library. The skill
\texttt{concise\_entity\_extraction} instructs the worker to extract the
shortest canonical proper noun and avoid unnecessary descriptive
clauses. Its procedural content remains unchanged; the failure instead
originates from missed or insufficiently broad activation. SPARO
therefore selects \textsc{RoutingEdit} rather than creating another
extraction skill.

\paragraph{Mutation.}
The optimizer keeps the skill procedure fixed and expands the observable
input patterns in its routing metadata.

\begin{quote}
\small
\textbf{Target skill.}
\texttt{concise\_entity\_extraction}\\
\textbf{Positive triggers before.}
\texttt{Who founded; What does ... stand for; this ``bubble'' burst;
president of the Senate; announced to Congress}\\
\textbf{Positive triggers after.}
\texttt{Who founded; What does ... stand for; this ``bubble'' burst;
president of the Senate; announced to Congress; Leonardo; da Vinci;
Burlington; Enoree; Long Cane; Honduras}\\

\end{quote}

\paragraph{Effect.}
The held-out trace records the corrected prediction
\texttt{Goodyear}. The iteration-12 routing-adjusted candidate is added
after its subsample score improves from 28 to 30 and obtains validation
relaxed exact match of 0.740. The same run reports
\texttt{Best skill test relaxed\_em: 0.761} for candidate 8. Overall,
the validation-selected program improves from 0.690 to 0.755 validation
relaxed exact match and from 0.735 to 0.751 held-out test relaxed exact
match.

\subsection{Case-Study Summary}
\label{app:case-summary}

The three cases illustrate distinct component responsibilities. Prompt
failures concern the global task contract or universal output policy,
such as returning a canonical answer span rather than an overly long
title. Skill failures concern missing or incorrect reusable procedures,
such as unit-scale validation in financial arithmetic. Routing failures
occur when useful existing knowledge is omitted from inputs where it
would help or activated on inputs where it does not apply. SPARO's key
behavior is therefore not simply to lengthen the prompt or accumulate
more skills; before generating a mutation, it uses controlled
counterfactual evidence to decide which component of the modular program
should be changed.